\documentclass[10pt,letterpaper,twocolumn]{article}
\usepackage[utf8]{inputenc}
\usepackage{amsmath}
\usepackage{amssymb}
\usepackage{graphicx}
\usepackage[unicode=true,
 bookmarks=false,
 breaklinks=true,pdfborder={0 0 1},backref=section,colorlinks=false]
 {hyperref}
\hypersetup{
 pagebackref=true,letterpaper=true,colorlinks}

\makeatletter

\pdfpageheight\paperheight
\pdfpagewidth\paperwidth

\providecommand{\tabularnewline}{\\}

\usepackage{cvpr}
\usepackage{times}
\usepackage{epsfig}
\usepackage{graphicx}
\usepackage{algorithm,algpseudocode}

\cvprfinalcopy 

\ifcvprfinal\fi

\makeatother

\begin{document}

\title{Regularized adversarial examples for model interpretability}

\author{Yoel Shoshan ~~~~~Vadim Ratner \\
 IBM Research\\
 IBM Haifa Labs, Haifa University Campus\\
 \texttt{\small{}yoels@il.ibm.com, vadimra@il.ibm.com}}
\maketitle
\begin{abstract}
As machine learning algorithms continue to improve, there is an increasing
need for explaining why a model produces a certain prediction for
a certain input. In recent years, several methods for model interpretability
have been developed, aiming to provide explanation of which subset
regions of the model input is the main reason for the model prediction.
In parallel, a significant research community effort is occurring
in recent years for developing adversarial example generation methods
for fooling models, while not altering the true label of the input,as
it would have been classified by a human annotator. In this paper,
we bridge the gap between adversarial example generation and model
interpretability, and introduce a modification to the adversarial
example generation process which encourages better interpretability.
We analyze the proposed method on a public medical imaging dataset,
both quantitatively and qualitatively, and show that it significantly
outperforms the leading known alternative method. Our suggested method
is simple to implement, and can be easily plugged into most common
adversarial example generation frameworks. Additionally, we propose
an explanation quality metric - $APE$ - ``Adversarial Perturbative
Explanation'', which measures how well an explanation describes model
decisions. 
\end{abstract}

\section{Introduction}

As machine learning algorithms continue to improve, there is an increasing
need for explaining why a given model produces a certain prediction.
The benefits of such explanation fall roughly into two categories:
explaining the end result for the user, and analysis of the network
by the researcher. An example of the former is providing localized
information on medical images classification to understand which areas
in a given image made a model to classify an image as malignant. An
example of the latter is that by seeing what caused decisions in a
model, one can understand what needs to be improved in the algorithm. 

In recent years, several methods for detecting saliency have been
developed, aiming to provide explanation of which subset regions of
the model input are the main reason for the model prediction. Part
of the methods \cite{ribeiro2016should,fong2017interpretable,chang2017interpreting}focus
on introducing perturbations into the input of a model and then analyzing
the modified model prediction.

In parallel, a significant effort is being put in recent years into
developing adversarial examples generation methods for fooling models,
usually aiming to keep the ``true label'' of the input, as will
be classified by a human reader (\cite{modas2018sparsefool,goodfellow2018explaining}).
The goal of such adversarial attacks is to identify unwanted behavior
of a network and exploit it. 

In this paper, we bridge the gap between the domains of network explanation
and adversarial examples. We introduce a modification to the adversarial
example generation process which aims to maximize interpretability.
Similarly to the domain of adversarial attacks, we change the input
in a way that affects the output of a model. Unlike previous works
on model explanation, our method does not require any reference ``deletion''
image and does not require training an additional NN model. Additionally,
our method provides explanation in the full resolution of the original
input.

We analyze the proposed method on a public medical imaging dataset,
both quantitatively and qualitatively, and show that it significantly
outperforms the leading known alternative method.

This modification is simple to implement, and can be easily plugged
in into most common adversarial example generation frameworks. The
resulting method is also applicable to non-image-related tasks.

The paper is organized as follows: Section \ref{sec:Related-Work}
reviews related work, in section \ref{sec:Proposed-Metric} we introduce
a new metric that we use to measure performance, section \ref{sec:Proposed-Method}
describes the proposed method, section \ref{sec:Experiments} presents
experimental data, section \ref{sec:Difference-from-existing} discusses
key differences from existing methods, and section \ref{sec:Conclusions}
discusses and concludes this work.


\section{\label{sec:Related-Work}Related Work}

When analyzing a model (e.g. neural network) based classifier, one
of the questions that arise is given an input and a classification
results, which parts of the input most affect the result. This gave
rise to several methods of deriving saliency maps. In particular,
several gradient-based techniques were proposed, such as \cite{simonyan2013deep},
computing the gradient of the classification loss with respect to
the image, and using the gradient magnitude as a measure of saliency.
However, this resulted in highly irregular saliency maps. This issue
was addressed by Fong and Vedaldi \cite{fong2017interpretable}. There,
the authors define perturbations of the input as a cross-fade between
the original image and an image that represents deletion of data.
Regularization terms were introduced when deriving the cross-fade
``mask'', which was then used as a saliency map. Chang $\etal$\cite{chang2017interpreting}
suggest an alternative method in which a generative model (Variational
Auto-Encoder) attempts to ``delete'' region by inpainting. Dabkowski
and Gal \cite{dabkowski2017real} achieve real-time performance by
training a network to directly generate saliency masks from images,
with the training data still dependent on mask and reference image.
Rey-de-Castro and Rabitz \cite{rey2018targeted} suggest adding a
perturbation generator network, a differentiable neural network component
that learns to distort the input image.

An important relevant field of study, which is gaining popularity
in recent years is adversarial attacks. \cite{goodfellow2018explaining}.
There, the original image is perturbed to modify a classifier prediction,
in a way that will not be noticed by a human reader. Usually, small
changes of the input are acceptable, as long as the ``true label''
(the label that a human will assign to this input) remains unchanged.

\section{\label{sec:Proposed-Metric}Proposed Metric}

\subsection{Saliency}

Defining image saliency in the context of a neural network is a non
trivial matter. In non formal terms, given a model, an input, and
a model prediction based on the given input, the saliency map should
explain how different parts of the input influenced the prediction.
There is strong motivation to find saliency generation methods and
metrics. A meaningful saliency map should help us see through the
“black box” of the model. It may help us discover situations when
a classifier model got the right answers but due to the wrong reasons.
For example, in the famous story in which a classifier was trained
to discover camouflaged tanks, and performed very well when looking
only at classification results. When the results were analyzed, it
was discovered that all of the camouflaged tanks images were taken
on cloudy days, while the tank-less images were taken on sunny days.
A proper saliency method would clearly have revealed this bias, and
show that the classifier didn't really learn to detect the tanks at
all. An additional example is explaining a medical imaging classifier
decision that helps to deliver more localized information on the decision,
which may help in providing more information to the human reader,
increase trust of the model and even lead to discovery of new features
undiscovered by humans until now. 

\subsection{Proposed APE metric - ``Adversarial Perturbative Explanation'' }

Ground truth (GT) based metrics, by themselves, do not represent well
explanation performance. Firstly, we are explaining a model prediction
behavior that may be flawed to begin with. In such case it is desirable
that the explanation will reveal the weakness of the model. Secondly,
there may be evidence outside the annotated GT object that may legitimately
influence a model decision. 

Therefore, to be able to quantitatively compare between explanation
methods, we formulate the $APE$ (Adversarial Perturbative Explanation)
metrics.

Perturbation based explanation describes regions that affect the decision
of a model, given a modified (perturbed) input. When creating a formal
metric, we considered two main aspects. Requirement 1: In the spirit
of Ockham's razor, we want the explanation to be as ``simple'' as
possible. Requirement 2: Additionally, it should suppress (or alternatively,
maintain) class evidence, depending whether a SDR (smallest destroying
region) or a SSR (smallest sufficient region) \cite{dabkowski2017real},
respectively, is required. We define two metrics corresponding to
SDR and SSR, $APE_{D}$ and $APE_{S}$ respectively.

Note: when deriving this metric we were inspired by Fong and Vedaldi
\cite{fong2017interpretable} loss formulation, with the key difference
that we use L0 directly and not an approximation of it in one of the
terms.

\subsubsection{$APE_{D}$ }

Let us consider a model $M$, an input $I$, a perturbed input $\tilde{I},$
and class index $i$. Firstly, we define $\hat{I}$ to be the clipped
version of $\tilde{I}$, constraining it to remain within valid input
value range. Let binarization be defined as:
\begin{equation}
B\left(f\right)\left(x\right)=\begin{cases}
1 & f\left(x\right)\neq0\\
0 & f\left(x\right)=0
\end{cases}\label{eq:8-1}
\end{equation}

we propose a saliency metric composed of the following terms:
\begin{enumerate}
\item Classification term: 
\begin{equation}
E_{c}\left(\hat{I}\right)=M\left(\hat{I}\right)_{i}\label{eq:1}
\end{equation}
$M(\hat{I})_{i}$being the prediction of the model given input $\hat{I}$
w.r.t. class index $i$. This term expresses the ``destructiveness''
of the saliency region w.r.t. class $i$. 
\item Sparsity term: 
\begin{equation}
E_{s}\left(I,\hat{I}\right)=L_{0}\left(I-\hat{I}\right)\label{eq:2}
\end{equation}
 
\item Smoothness term: 
\begin{equation}
E_{r}\left(I,\hat{I}\right)=TV\left(B\left(I-\hat{I}\right)\right)\label{eq:3}
\end{equation}
\end{enumerate}
The overall proposed saliency metric is defined as follows:
\begin{equation}
\frac{1}{N}L_{0}\left(I-\hat{I}\right)+\frac{1}{N}TV\left(B\left(I-\hat{I}\right)\right)+M\left(\hat{I}\right)_{i}\label{eq:4}
\end{equation}

$N$ = number of elements on $I$ 

Classification term Eq.\ref{eq:1} addresses requirement 1. Sparsity
and smoothness terms Eqs. \ref{eq:2}, \ref{eq:3}(combined) address
requirement 2.

\subsubsection{$APE_{S}$ }

Similarly to the SDR version, we define $\hat{I}$ to be the clipped
version of $\tilde{I}$, constraining it to remain within valid input
value range. 
\begin{enumerate}
\item Classification term: 
\begin{equation}
E_{c}\left(\hat{I}\right)=\left|M\left(\hat{I}\right)_{i}-M(I)_{i}\right|\label{eq:1-1}
\end{equation}
$M(\hat{I})_{i}$being the prediction of the model given input $\hat{I}$
w.r.t. class index $i$. This term expresses how well the original
model classification is preserved. 
\item Sparsity term: 
\begin{equation}
E_{s}\left(I,\hat{I}\right)=L_{0}\left(1-B\left(I-\hat{I}\right)\right)\label{eq:2-1}
\end{equation}
 
\item Smoothness term: 
\begin{equation}
E_{r}\left(I,\hat{I}\right)=TV\left(1-B\left(I-\hat{I}\right)\right)\label{eq:3-1}
\end{equation}
\end{enumerate}
The overall proposed saliency metric is defined as follows:
\begin{multline}
\frac{1}{N}L_{0}\left(1-B\left(I-\hat{I}\right)\right)+\frac{1}{N}TV\left(1-B\left(I-\hat{I}\right)\right)+\\
\left|M\left(\hat{I}\right)_{i}-M(I)_{i}\right|.\label{eq:4-1}
\end{multline}

$N$ = number of elements on $I$ 

\subsubsection{$APE$ terms weights}

In some settings, it may prove useful to define coefficients that
provide different weights to each term. 

We formalize such term weighting in eq. \ref{eq:4-2}, $\alpha_{sp}$,
$\alpha_{sm}$ and $\alpha_{cl}$ being the sparsity, smoothness and
classification weight respectively. We formalize it for $APE_{D}$,
but it can be symmetrically formalized for $APE_{S}$which we omit
for brevity.

\begin{equation}
\alpha_{sp}\frac{1}{N}L_{0}\left(I-\hat{I}\right)+\alpha_{sm}\frac{1}{N}TV\left(B\left(I-\hat{I}\right)\right)+\alpha_{cl}M\left(\hat{I}\right)_{i}\label{eq:4-2}
\end{equation}

In this paper, when measuring $APE$ based performance (table \ref{tab:MP-min-=00003D}),
we only use $\alpha_{sp}=1,\alpha_{sm}=1,\alpha_{cl}=1$ (eq. \ref{eq:4-2})
as we feel that this is appropriate for the examined domain and task.
However, it is possible that in a different domain or task, different
metric coefficients will be more suitable. For example, if in the
examined domain or task, it is known in advance that objects are relatively
big, $\alpha_{sm}$ can be increased and possibly $\alpha_{sp}$ can
be reduced. Such modification will strongly favor smooth connected
components explanations and discourage too sparse explanations.

\section{\label{sec:Proposed-Method}Proposed Method}

We focus only on optimizing w.r.t. $APE_{D}$ , as we believe that
in certain domains, such as medical imaging, $SDR$ is preferred,
since it explores less drastic modifications to the input and allows
the model to consider larger context.

Let $\hat{I}$ be the perturbed input. Since support function is not
continuously differentiable, we approximate binarization Eq. \ref{eq:8-1}
by the following:
\begin{multline}
B\left(I-\hat{I}\right)\left(x\right)\eqsim S\left(x\right)\equiv\\
\frac{2}{1+exp\left(-\gamma\left(\left|I\left(x\right)-\hat{I}\left(x\right)\right|-\varepsilon\right)\right)}-1.\label{eq:6}
\end{multline}
The result approximates a smoothed step function which receives negative
values below $\varepsilon$ (we found $\gamma=30,\varepsilon=0.01$
to achieve good results).

Inspired by Kurakin $\etal$\cite{kurakin2016adversarial} we formulate
an optimization problem which aims to find $\hat{I}$ that minimizes
the proposed $APE_{D}$ metric \ref{eq:4}. 

\textbf{Phase 1:}We initialize $\hat{I}$ to $I$. Then, we iteratively
modify $\hat{I}$ using the gradient of the loss w.r.t. the input,
while keeping model $M$ frozen. 

Any gradient based optimizer may be used, including, for example,
SGD \cite{robbins1951}, Adam \cite{kingma2014adam}, AdaDelta \cite{zeiler2012adadelta}.
For brevity we describe the SGD update step, while in practice we
use Adam \cite{kingma2014adam} optimizer.. 

\begin{equation}
\triangle\hat{I}=-\delta\frac{\partial E}{\partial\hat{I}_{k}}\label{eq:5}
\end{equation}

The loss $E$ is a smooth version of the saliency metric:

\begin{equation}
E\left(I,\hat{I}\right)=M\left(\hat{I}\right)_{i}+\alpha sum\left(S\right)+\beta TV\left(S\right)\label{eq:7}
\end{equation}

The first term reduces classification value of class $i$. The second
term approximates the size of support of $P$, with two important
differences. Firstly, as was mentioned previously, it is smooth w.r.t.
$S$. Secondly, very small values of the perturbation result in negative
values of $S$, decreasing the overall value of the second member.
This should encourage close-to-zero perturbations over most of the
image. This will also be useful later when we derive the saliency
mask. The third term encourages smoothness of $S$, preferring continues
regions of non-zero values over scattered individual elements (pixels
in the case of images).

On each iteration, after computing the update step Eq. \ref{eq:5}we
constrain $\hat{I}$ to remain in the original applicable values range
which $I$ is sampled from, by clipping its values. After either completing
a defined number of iterations, or a when reaching convergence, we
derive the saliency mask by thresholding $S$ at zero:

\begin{equation}
sal\left(x\right)=\begin{cases}
1 & S\left(x\right)\geq0\\
0 & S\left(x\right)<0
\end{cases}\label{eq:8}
\end{equation}

We found that this is sufficient for the purpose of explanation, however,
zeroing out some regions of $P$ may increase the classification term,
increasing the overall $APE_{D}$score. We therefore introduce phase
2, which finds the smallest achievable classification probability
(for class $i$), given that perturbations are only allowed within
the mask derived in \ref{eq:8}

\textbf{Phase 2:} Small (below $\varepsilon$) perturbations of the
input may also affect classification (and classification term), and
eliminating those perturbations in Eq. \ref{eq:8} may increase overall
loss. In order to guarantee that classification loss remain small
for the derived saliency mask, we introduce the second phase of the
algorithm. The purpose of this stage is to make sure that we minimize
the classification term while allowing perturbations only within the
mask derived in Eq. \ref{eq:8}. For this purpose we find a perturbed
image, similarly to Eq. \ref{eq:5}, that is non-zero only inside
the mask, starting with $\check{I}=I$,:
\begin{equation}
\triangle\check{I}\left(x\right)=-\delta\frac{\partial E}{\partial\check{I}_{k}}sal\left(x\right)\label{eq:9}
\end{equation}

This time, we only minimize the classification term.

\begin{equation}
E\left(\check{I}\right)=M\left(\check{I}\right)_{i}\label{eq:10}
\end{equation}

On this second phase we drop both sparsity and smoothness terms to
allow non-regulated changes to occur within the mask regions, compensating
for ommiting the out-of-mask changes.

\section{\label{sec:Experiments}Experiments}

\begin{figure*}[t]
\centering{}%
\begin{minipage}[t][0.18\paperheight]{0.28\paperwidth}%
\begin{flushleft}
\includegraphics[clip,width=0.2\paperwidth,height=0.15\paperheight]{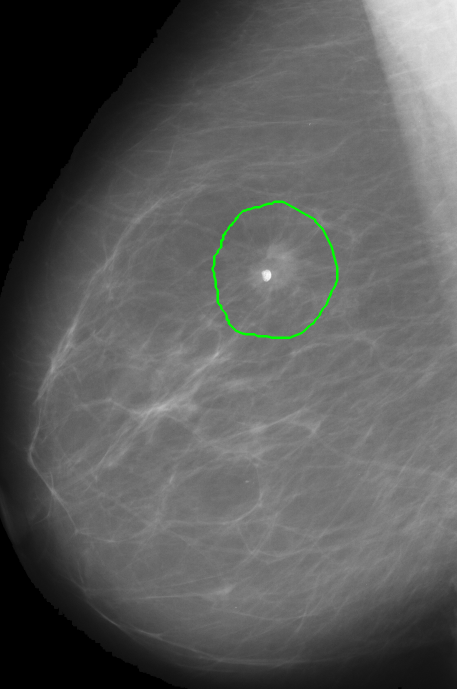}
\par\end{flushleft}%
\end{minipage}%
\begin{minipage}[t][0.18\paperheight]{0.28\paperwidth}%
\begin{flushleft}
\includegraphics[bb=0bp 0bp 460bp 688bp,clip,width=0.2\paperwidth,height=0.15\paperheight]{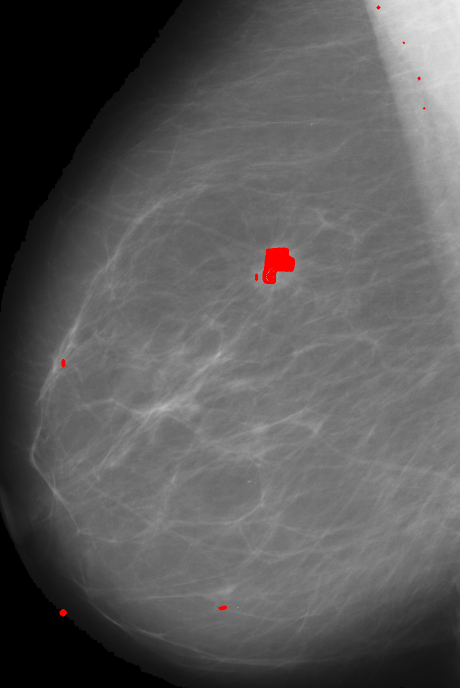}
\par\end{flushleft}%
\end{minipage}%
\begin{minipage}[t][0.18\paperheight]{0.28\paperwidth}%
\begin{flushleft}
\includegraphics[bb=0bp 0bp 458bp 691bp,clip,width=0.2\paperwidth,height=0.15\paperheight]{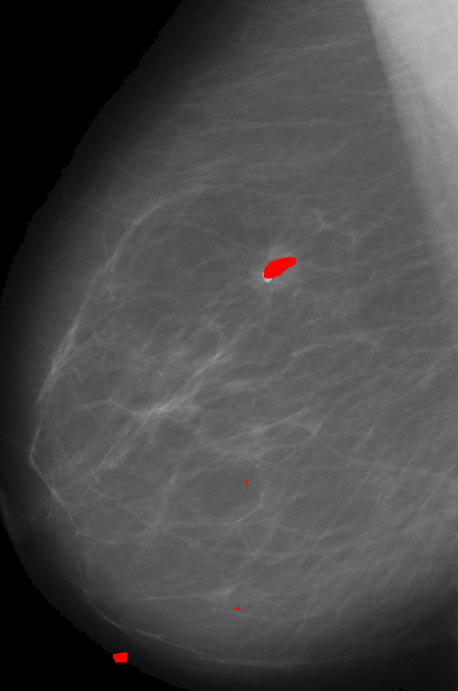}
\par\end{flushleft}%
\end{minipage}\\
\begin{minipage}[t][0.18\paperheight]{0.28\paperwidth}%
\begin{flushleft}
\includegraphics[bb=0bp 0bp 451bp 560bp,clip,width=0.2\paperwidth,height=0.15\paperheight]{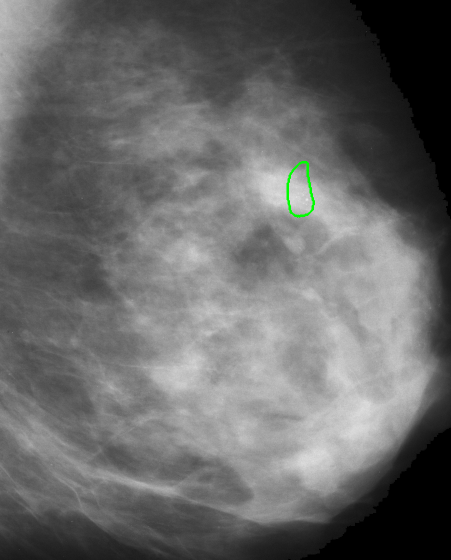}
\par\end{flushleft}%
\end{minipage}%
\begin{minipage}[t][0.18\paperheight]{0.28\paperwidth}%
\begin{flushleft}
\includegraphics[bb=0bp 0bp 443bp 560bp,clip,width=0.2\paperwidth,height=0.15\paperheight]{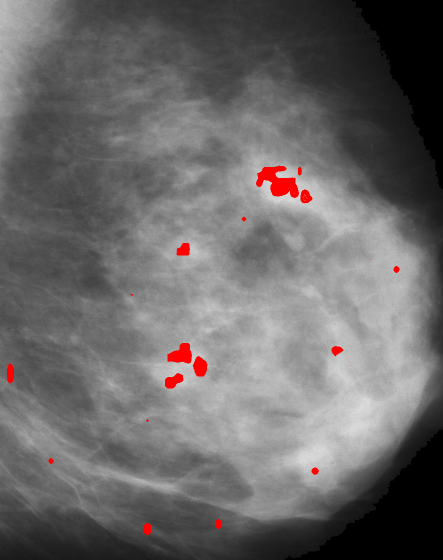}
\par\end{flushleft}%
\end{minipage}%
\begin{minipage}[t][0.18\paperheight]{0.28\paperwidth}%
\begin{flushleft}
\includegraphics[bb=0bp 0bp 449bp 560bp,clip,width=0.2\paperwidth,height=0.15\paperheight]{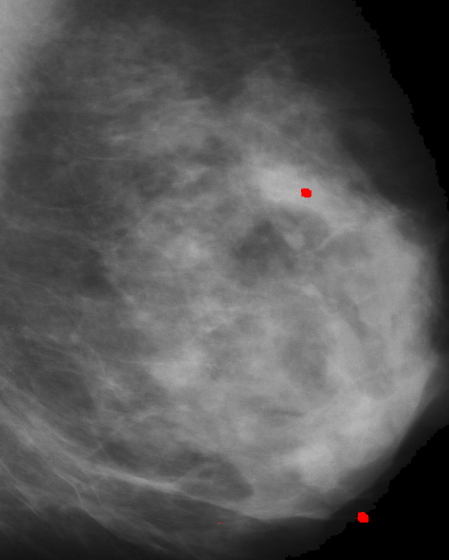}
\par\end{flushleft}%
\end{minipage}\\
\begin{minipage}[t][0.18\paperheight]{0.28\paperwidth}%
\begin{flushleft}
\includegraphics[bb=0bp 0bp 670bp 880bp,clip,width=0.2\paperwidth,height=0.15\paperheight]{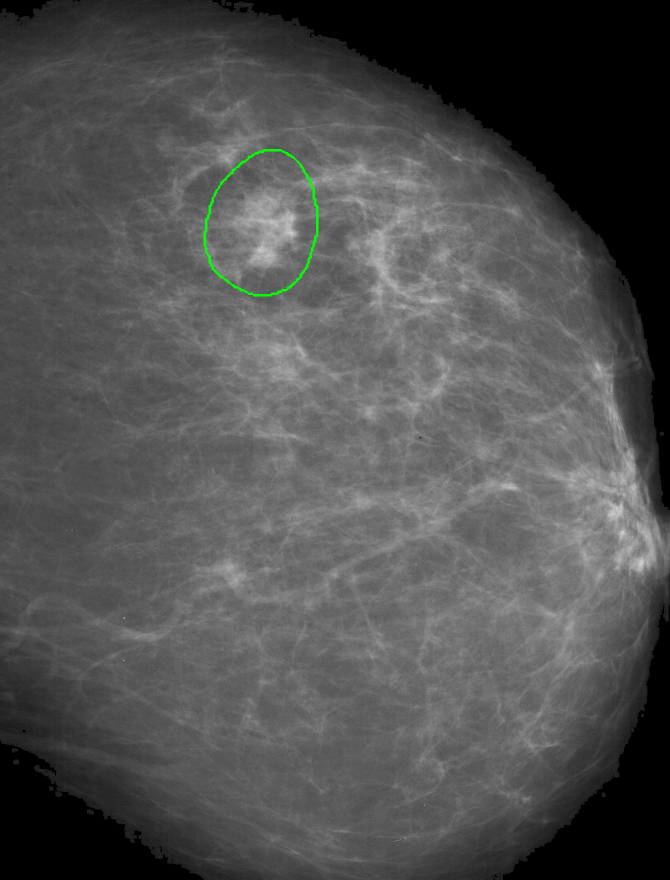}
\par\end{flushleft}%
\end{minipage}%
\begin{minipage}[t][0.18\paperheight]{0.28\paperwidth}%
\begin{flushleft}
\includegraphics[bb=0bp 0bp 671bp 881bp,clip,width=0.2\paperwidth,height=0.15\paperheight]{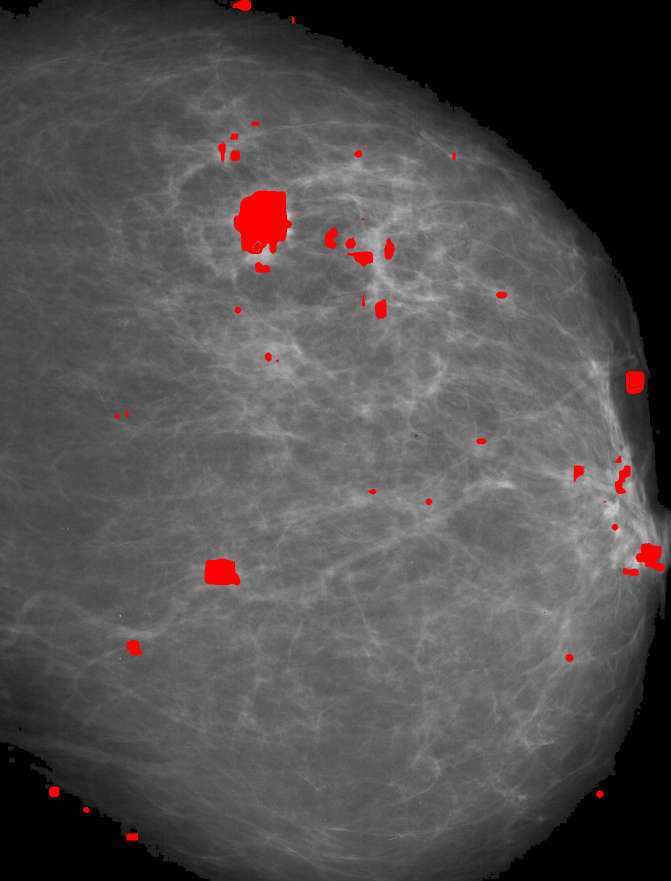}
\par\end{flushleft}%
\end{minipage}%
\begin{minipage}[t][0.18\paperheight]{0.28\paperwidth}%
\begin{flushleft}
\includegraphics[bb=0bp 0bp 671bp 880bp,clip,width=0.2\paperwidth,height=0.15\paperheight]{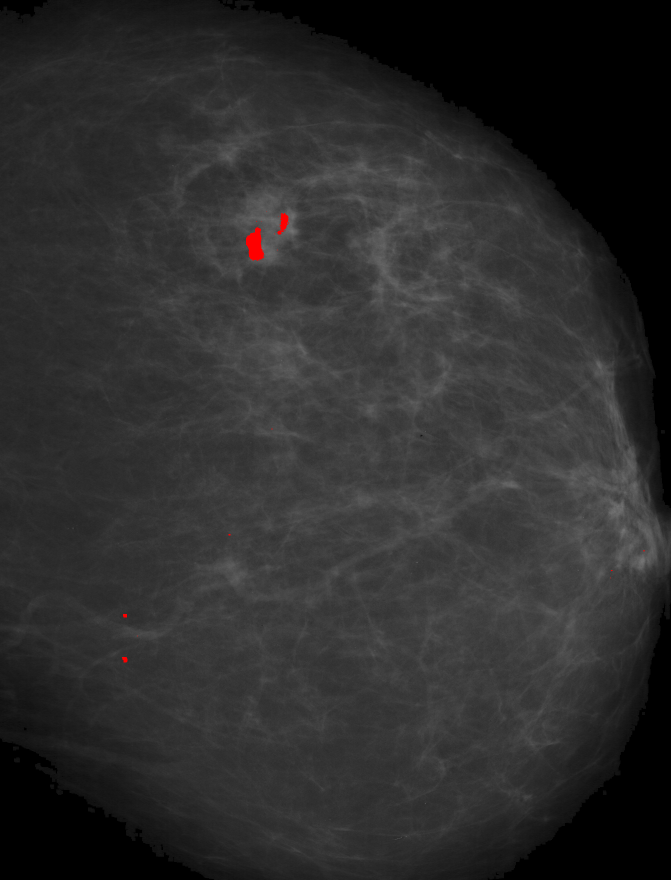}
\par\end{flushleft}%
\end{minipage}\\
\begin{minipage}[t][0.18\paperheight]{0.28\paperwidth}%
\begin{flushleft}
\includegraphics[bb=0bp 0bp 340bp 860bp,clip,width=0.2\paperwidth,height=0.15\paperheight]{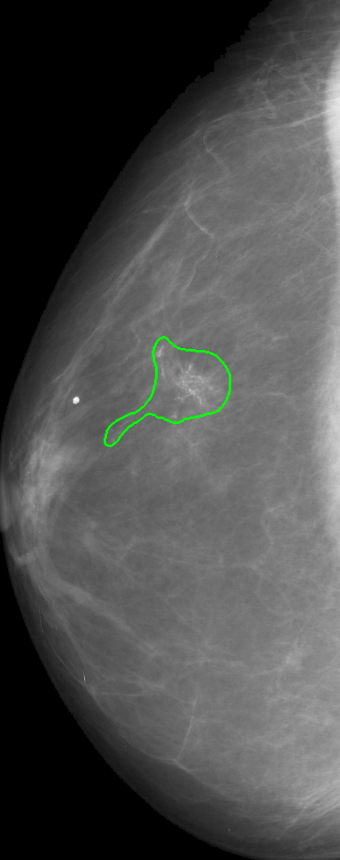}
\par\end{flushleft}%
\end{minipage}%
\begin{minipage}[t][0.18\paperheight]{0.28\paperwidth}%
\begin{flushleft}
\includegraphics[bb=0bp 0bp 340bp 860bp,clip,width=0.2\paperwidth,height=0.15\paperheight]{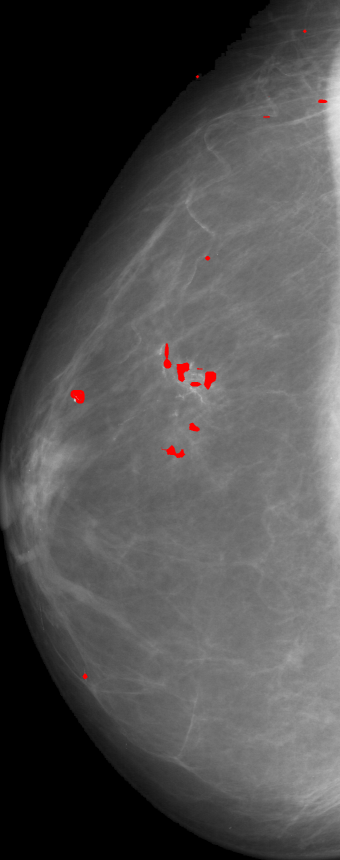}
\par\end{flushleft}%
\end{minipage}%
\begin{minipage}[t][0.18\paperheight]{0.28\paperwidth}%
\begin{flushleft}
\includegraphics[clip,width=0.2\paperwidth,height=0.15\paperheight]{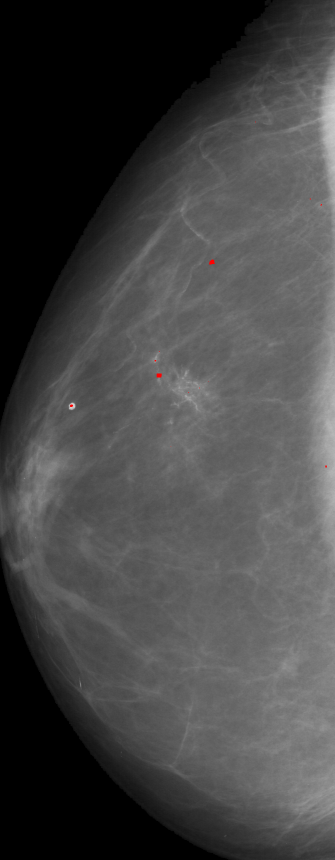}
\par\end{flushleft}%
\end{minipage}\\
\begin{minipage}[t][0.03\paperheight]{0.24\paperwidth}%
\begin{center}
a
\par\end{center}%
\end{minipage}~%
\begin{minipage}[t][0.03\paperheight]{0.26\paperwidth}%
\begin{center}
b
\par\end{center}%
\end{minipage}~%
\begin{minipage}[t][0.03\paperheight]{0.26\paperwidth}%
\begin{center}
c 
\par\end{center}%
\end{minipage}\caption{\label{fig:1}Column a: ground truth with malignant lesion delineated
in green, column b: meaningful perturbations mask \cite{fong2017interpretable},
column c: proposed method resulting mask. As can be seen in table
\ref{tab:MP-min-=00003D}, the proposed method changes much smaller
parts of the input, while reaching better classification term. }
\end{figure*}

We compare our method with Fong and Vedaldi (\cite{fong2017interpretable})
both qualitatively and quantitatively. Quantitative comparison is
based on two metrics. One is the metric discussed in section \ref{sec:Proposed-Metric}
($APE_{D}$). The other measures a ground truth (GT) match by counting
the portion of the explanation mask CCs (Connected Components) that
intersect with the tested object. The proposed method performs significantly
better performance w.r.t. both metrics (Table \ref{tab:MP-min-=00003D}).

To evaluate our method we have selected the DDSM dataset \cite{heath2000digital},
a digital database for screening mammography. The main reasons are
a. The dataset is publicly available; b. The images are high resolution
averaging at around 6000x4000 pixels; c. The objects are of varying
sizes ranging between pixel sized micro-calcifications to large tumors;
d. Malignancy classification is a hard and non-trival task, representing
an interesting setting to explore explainability.

\subsection{Model and experiment setting}

For the purpose of the experiment we train a single inception-resnet-v3\cite{szegedy2016rethinking}
model on per-image classification task of malignant vs. non-malignant
images. The model architecture is modified to accept single-channel
inputs (grayscale). Feature extraction layers up to 2 layers after
``mixed\_6a'' layer are kept, followed by a fully connected layer
of size 256, and finally a fully connected layer of size 2, followed
by a softmax operation over the two classes ``malignant'' vs. ``non-malignant''.
It is important to note that, during training, the model is never
exposed to localization information, and is trained w.r.t. the overall
image malignancy label (the presence of at least a single malignant
finding). This results in \textasciitilde{}8M trainable parameters.
Standard batch-normalization was used. 

We randomly split the dataset into 80\% train-set (\textasciitilde{}8000
images) and 20\% validation-set (\textasciitilde{}2000 images), making
sure that no patient appears on both train and validation sets. The
model is tested on the validation set on which it achieves a \textasciitilde{}0.8
rocauc in the mentioned classification task. The qualitative and quantitative
(table \ref{tab:MP-min-=00003D}) results are calculated on the validation
set.

We explore several setups (Table \ref{tab:MP-min-=00003D}) of both
methods w.r.t. $L_{0}$approximation, Total Variation (TV) and $tv_{\gamma}$
(Total Variation $\gamma$) \cite{fong2017interpretable}. Masks for
Fong and Vedaldi \cite{fong2017interpretable} are thresholded with
T value which provides the best $APE_{D}$score for the entire validation
set, scanned at steps of 1e-6. Masks for our proposed method are calculated
by thresholding with 0 (eq. \ref{eq:8}) $\gamma=30,\varepsilon=0.01$
(eq. \ref{eq:6}). We do not scan for additional $\gamma$and $\epsilon$values
as our initial ``guess'' worked well.

\subsection{GT localization metric}

DDSM dataset contains localized GT (ground truth) information delineating
malignant lesions. Since it is important to know if the perturbation
method managed to ``fool'' the model in nonsensical ways, especially
in the context of adversarial example generation, we examine how well
the explanation masks correlate with the actual findings in the images.
For this purpose, we take each generated mask, and measure which percentage
of its CCs (connected components) have non zero intersection with
GT lesions/objects. See table \ref{tab:MP-min-=00003D}. 

\begin{figure*}[t]
\centering{}%
\begin{minipage}[t][0.14\paperheight]{0.2\paperwidth}%
\begin{flushleft}
\includegraphics[bb=170bp 350bp 325bp 450bp,clip,width=0.17\paperwidth,height=0.13\paperheight]{00463_gt_contours}
\par\end{flushleft}%
\end{minipage}%
\begin{minipage}[t][0.14\paperheight]{0.2\paperwidth}%
\begin{flushleft}
\includegraphics[bb=245bp 530bp 400bp 630bp,clip,width=0.17\paperwidth,height=0.13\paperheight]{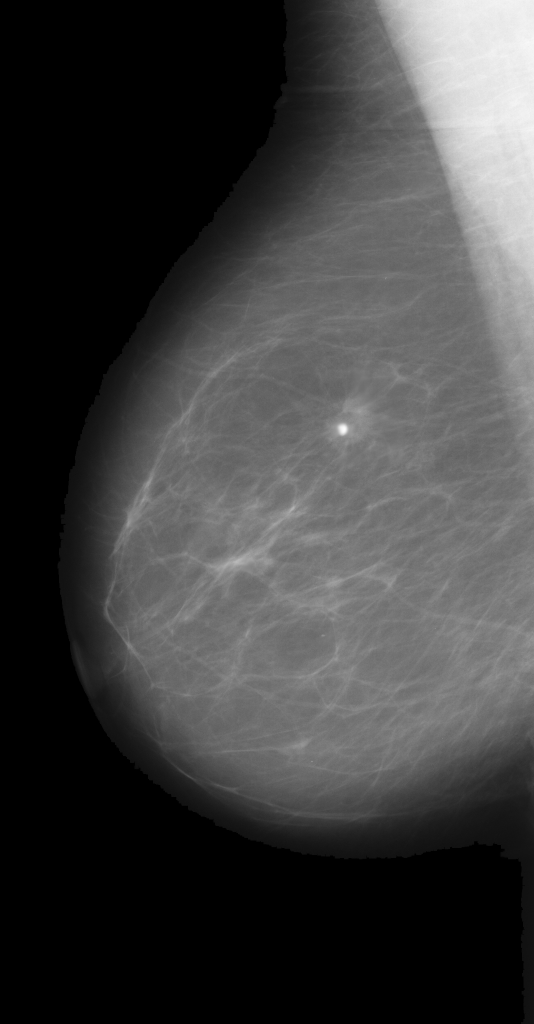}
\par\end{flushleft}%
\end{minipage}%
\begin{minipage}[t][0.14\paperheight]{0.2\paperwidth}%
\begin{flushleft}
\includegraphics[bb=245bp 530bp 400bp 630bp,clip,width=0.17\paperwidth,height=0.13\paperheight]{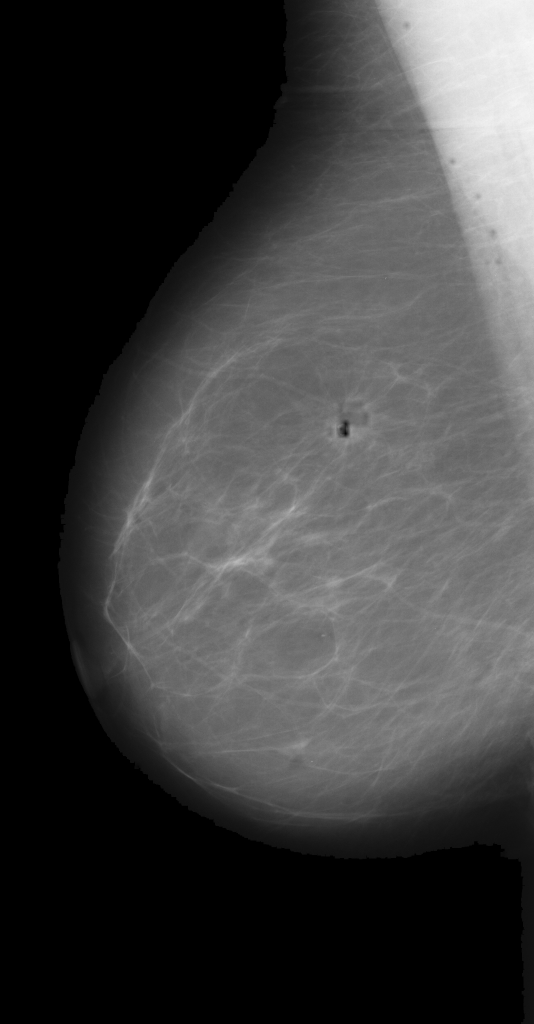}
\par\end{flushleft}%
\end{minipage}%
\begin{minipage}[t][0.14\paperheight]{0.2\paperwidth}%
\begin{flushleft}
\includegraphics[bb=170bp 350bp 325bp 450bp,clip,width=0.17\paperwidth,height=0.13\paperheight]{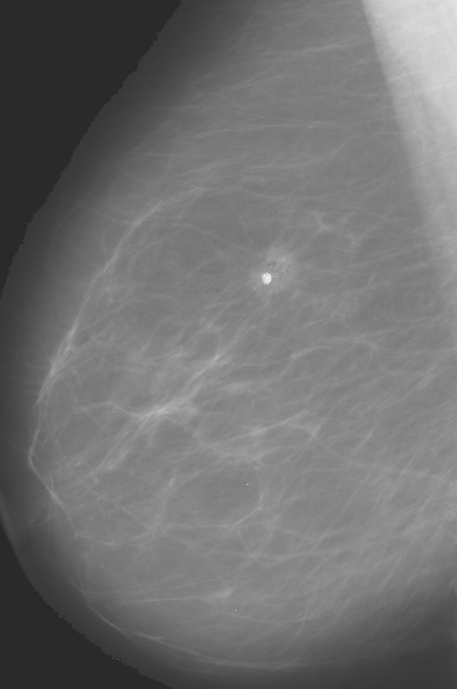}
\par\end{flushleft}%
\end{minipage}\\
\begin{minipage}[t][0.14\paperheight]{0.2\paperwidth}%
\begin{flushleft}
\includegraphics[bb=200bp 300bp 350bp 425bp,clip,width=0.17\paperwidth,height=0.13\paperheight]{00599_gt_contours}
\par\end{flushleft}%
\end{minipage}%
\begin{minipage}[t][0.14\paperheight]{0.2\paperwidth}%
\begin{flushleft}
\includegraphics[bb=200bp 400bp 350bp 525bp,clip,width=0.17\paperwidth,height=0.13\paperheight]{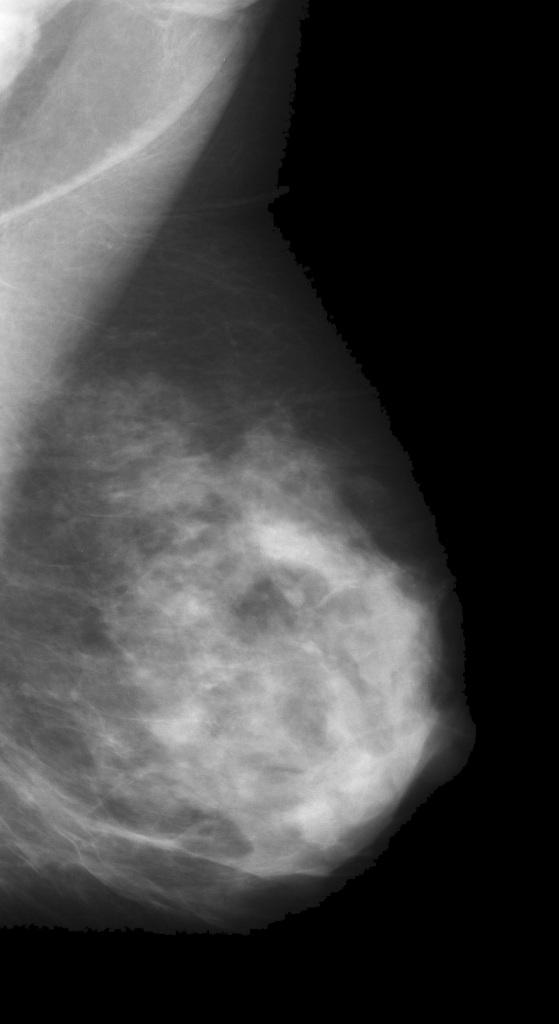}
\par\end{flushleft}%
\end{minipage}%
\begin{minipage}[t][0.14\paperheight]{0.2\paperwidth}%
\begin{flushleft}
\includegraphics[bb=200bp 400bp 350bp 525bp,clip,width=0.17\paperwidth,height=0.13\paperheight]{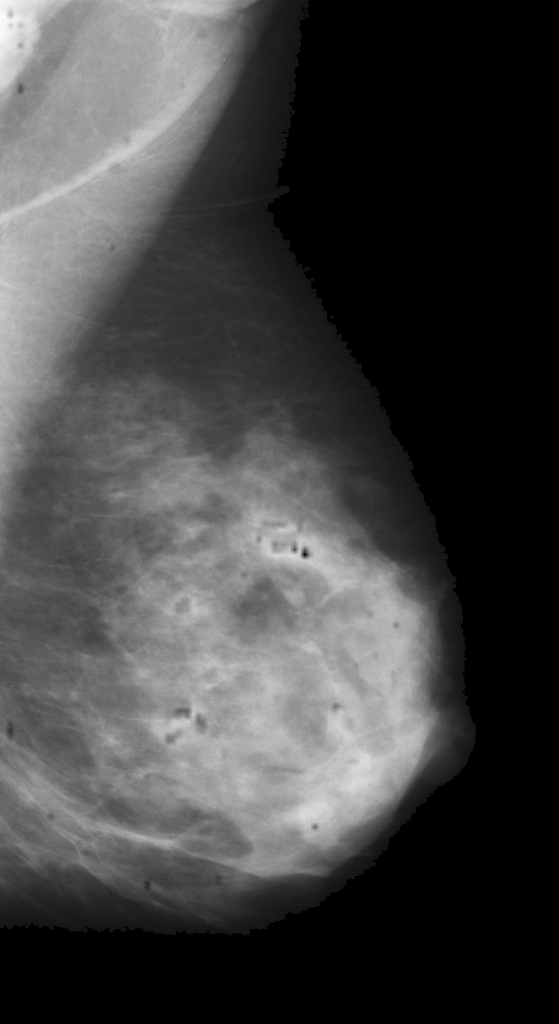}
\par\end{flushleft}%
\end{minipage}%
\begin{minipage}[t][0.14\paperheight]{0.2\paperwidth}%
\begin{flushleft}
\includegraphics[bb=200bp 300bp 350bp 425bp,clip,width=0.17\paperwidth,height=0.13\paperheight]{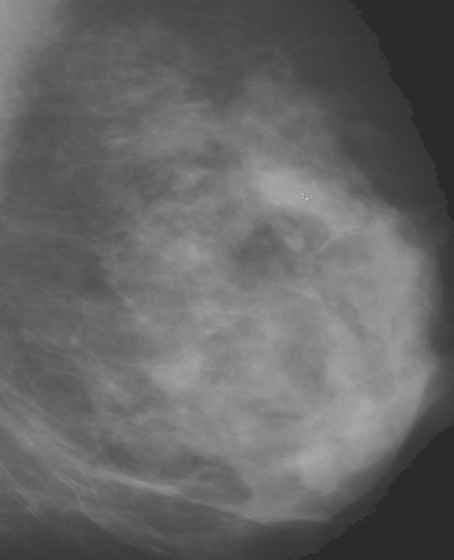}
\par\end{flushleft}%
\end{minipage}\\
\begin{minipage}[t][0.14\paperheight]{0.2\paperwidth}%
\begin{flushleft}
\includegraphics[bb=100bp 550bp 400bp 800bp,clip,width=0.17\paperwidth,height=0.13\paperheight]{00997_gt_contours}
\par\end{flushleft}%
\end{minipage}%
\begin{minipage}[t][0.14\paperheight]{0.2\paperwidth}%
\begin{flushleft}
\includegraphics[bb=150bp 650bp 450bp 900bp,clip,width=0.17\paperwidth,height=0.13\paperheight]{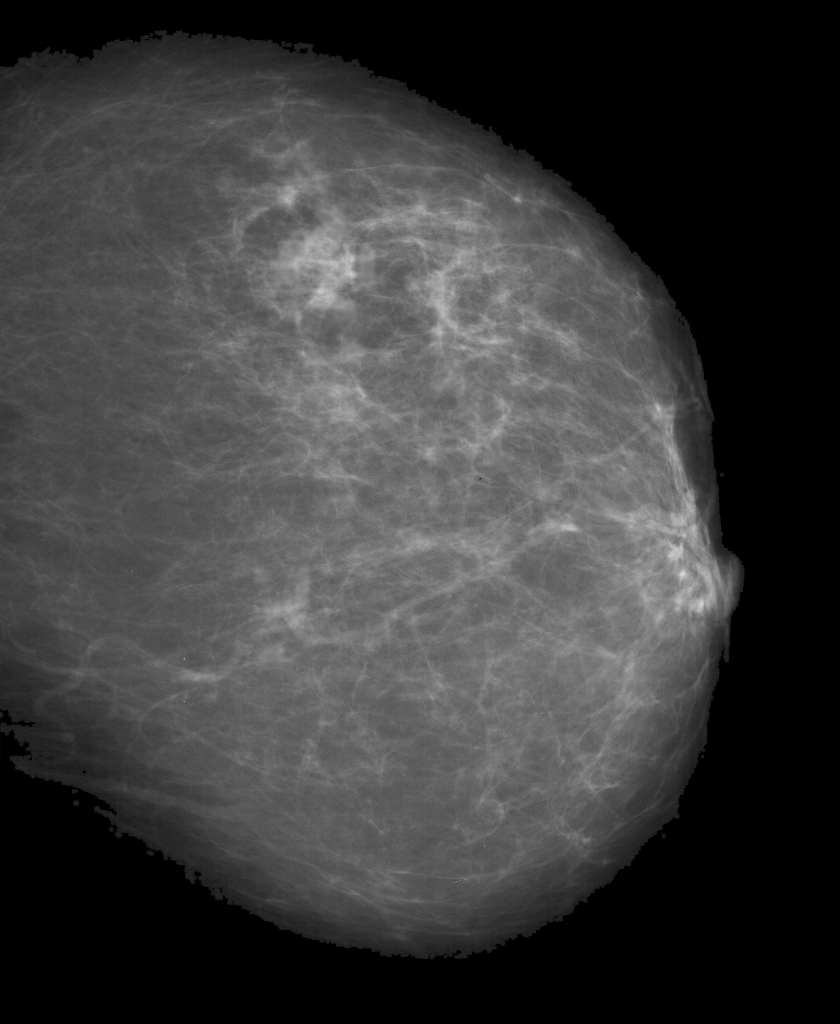}
\par\end{flushleft}%
\end{minipage}%
\begin{minipage}[t][0.14\paperheight]{0.2\paperwidth}%
\begin{flushleft}
\includegraphics[bb=150bp 650bp 450bp 900bp,clip,width=0.17\paperwidth,height=0.13\paperheight]{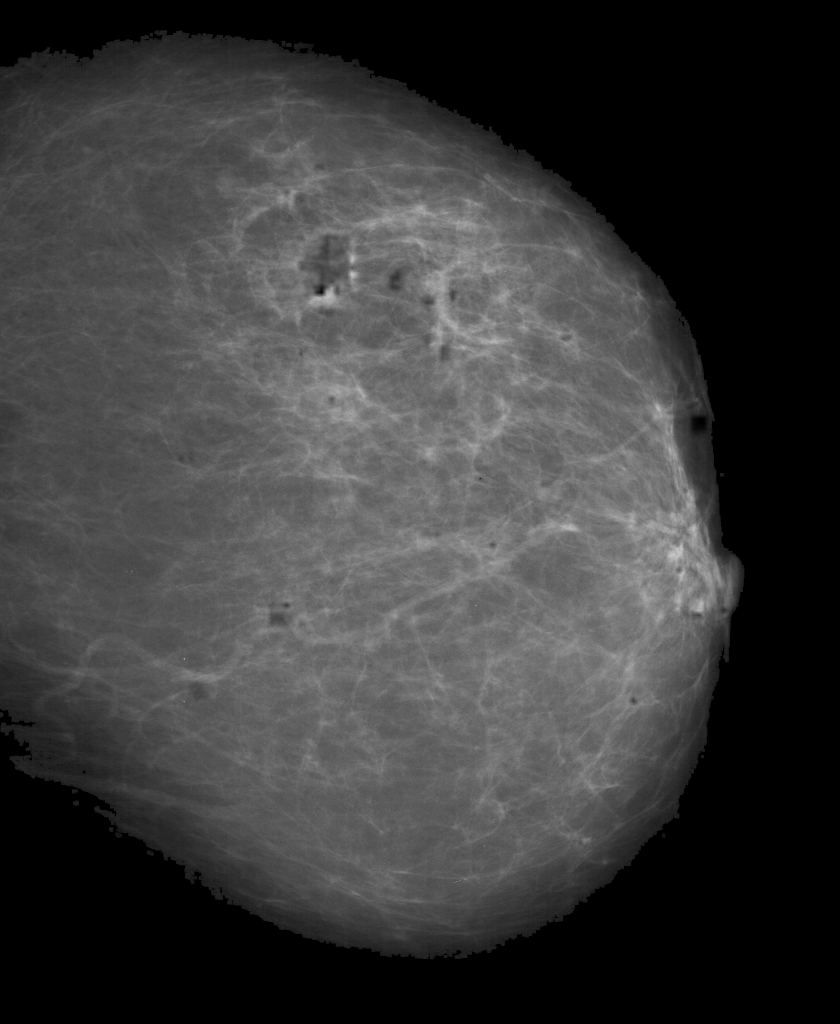}
\par\end{flushleft}%
\end{minipage}%
\begin{minipage}[t][0.14\paperheight]{0.2\paperwidth}%
\begin{flushleft}
\includegraphics[bb=100bp 550bp 400bp 800bp,clip,width=0.17\paperwidth,height=0.13\paperheight]{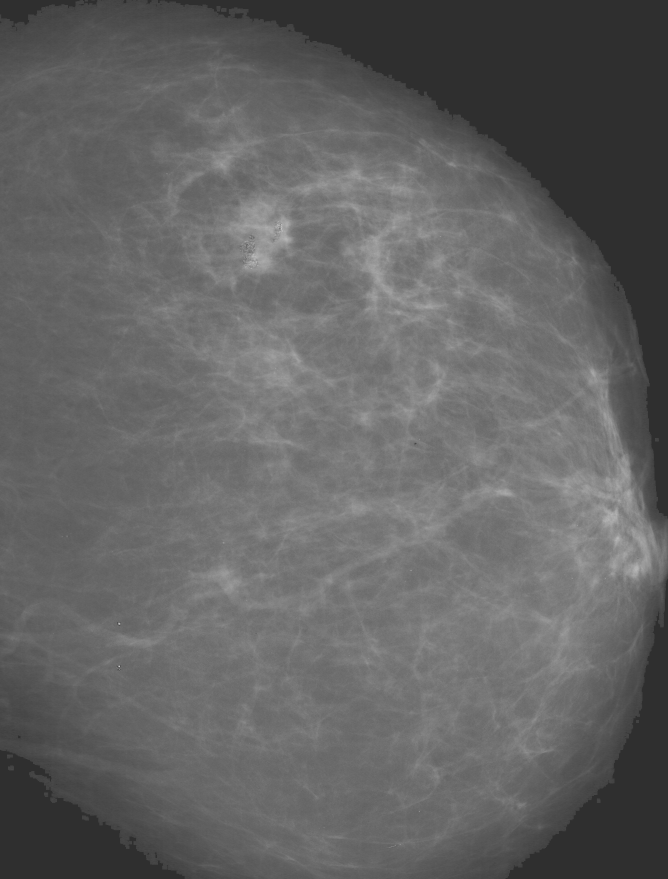}
\par\end{flushleft}%
\end{minipage}\\
\begin{minipage}[t][0.14\paperheight]{0.2\paperwidth}%
\begin{flushleft}
\includegraphics[bb=60bp 400bp 225bp 600bp,clip,width=0.17\paperwidth,height=0.13\paperheight]{00740_gt_contours}
\par\end{flushleft}%
\end{minipage}%
\begin{minipage}[t][0.14\paperheight]{0.2\paperwidth}%
\begin{flushleft}
\includegraphics[bb=60bp 475bp 225bp 675bp,clip,width=0.17\paperwidth,height=0.13\paperheight]{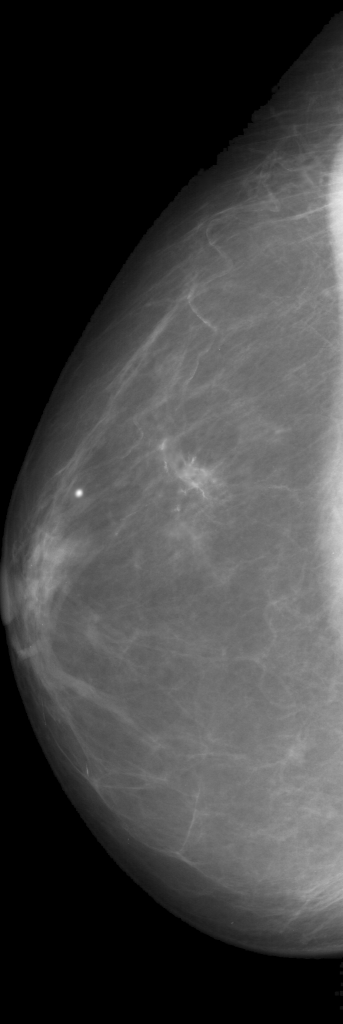}
\par\end{flushleft}%
\end{minipage}%
\begin{minipage}[t][0.14\paperheight]{0.2\paperwidth}%
\begin{flushleft}
\includegraphics[bb=60bp 475bp 225bp 675bp,clip,width=0.17\paperwidth,height=0.13\paperheight]{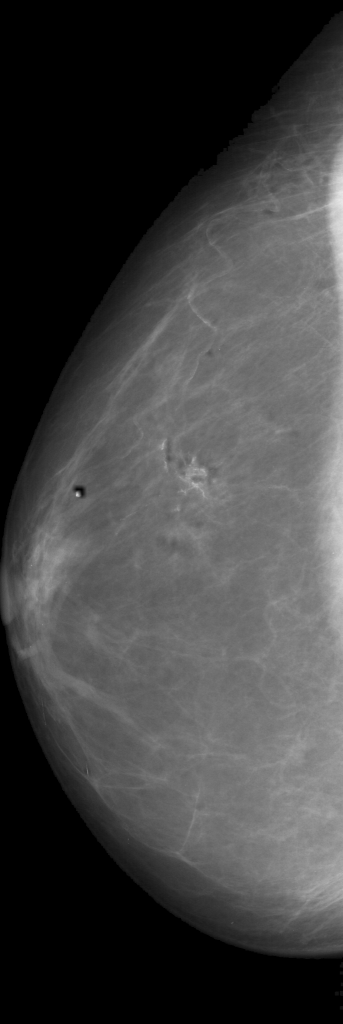}
\par\end{flushleft}%
\end{minipage}%
\begin{minipage}[t][0.14\paperheight]{0.2\paperwidth}%
\begin{flushleft}
\includegraphics[bb=60bp 400bp 225bp 600bp,clip,width=0.17\paperwidth,height=0.13\paperheight]{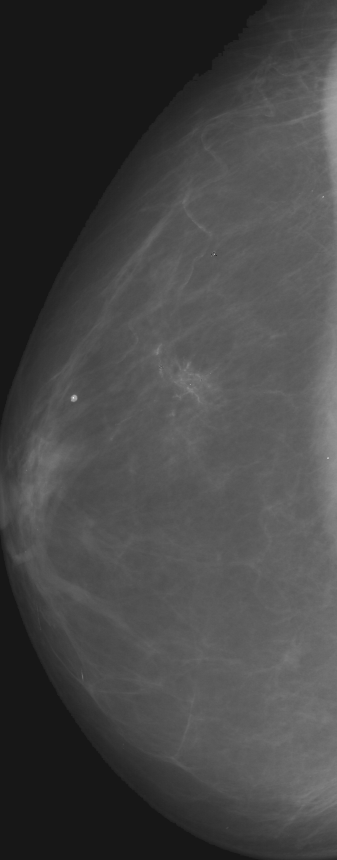}
\par\end{flushleft}%
\end{minipage}\\
\begin{minipage}[t][0.03\paperheight]{0.18\paperwidth}%
\begin{center}
a
\par\end{center}%
\end{minipage}~%
\begin{minipage}[t][0.03\paperheight]{0.18\paperwidth}%
\begin{center}
b
\par\end{center}%
\end{minipage}~%
\begin{minipage}[t][0.03\paperheight]{0.18\paperwidth}%
\begin{center}
c
\par\end{center}%
\end{minipage}~%
\begin{minipage}[t][0.03\paperheight]{0.18\paperwidth}%
\begin{center}
d 
\par\end{center}%
\end{minipage}\caption{\label{fig:2}Column a: ground truth with malignant lesion delineated
in green, column b: meaningful perturbations \cite{fong2017interpretable}
smoothing perturbation, column c: meaningful perturbations \cite{fong2017interpretable}
reduction to zero perturbation, column d: proposed method adversarial
perturbation. It is interesting to note that while our method generates
adversarial noise-like patterns, it is mostly formed within well localized
adversarial ``patch''.}
\end{figure*}

\subsection{Results}

As demonstrated in Table \ref{tab:MP-min-=00003D}, our method achieves
best performance on both $APE_{D}$ and $GT$ localization metrics.
Furthermore, we achieve top results on each separate $APE_{D}$component.
Figure 1 \ref{fig:1} shows 4 representative example images, demonstrating
GT malignant lesions, our proposed method explanation mask and an
alternative method explanation mask. Figure 2 \ref{fig:2}shows how
the adversarial perturbation images look like for the compared methods,
zoomed in at the malignant lesion (the object that the model should
search for). It can be seen how Fong and Vedaldi \cite{fong2017interpretable}
either darkens or blurs regions localized at the lesion area, depending
on the reference deletion image, while our proposed method generates
localized adversarial ``patches''. It can also be seen that our
method generates significantly less localized false positive mask
locations when compared with the GT, which explains why our proposed
method got, quantitatively, better GT localization performance results.
(``GT localization results'' column in table \ref{tab:MP-min-=00003D}).

\begin{table*}
\begin{centering}
\begin{tabular}{|c|c|c|c|c|c|c|c|}
\hline 
\multicolumn{4}{|c|}{experiment setting} & \multicolumn{4}{c|}{$APE_{D}$results ($\alpha_{sm}=1,\alpha_{sp}=1,\alpha_{cl}=1)$}\tabularnewline
\hline 
\hline 
Method & $L_{0}$approx. coeff & tv coeff & $tv_{\gamma}$ & $L_{0}$ & tv & classification & $APE_{D}$ \tabularnewline
\hline 
MP min & 0.01 & 0.2 & 1 & 0.9947 & 0.0019 & 0.0044 & 1.001\tabularnewline
\hline 
MP min & 0.01 & 0.2 & 3 & 0.5254 & 0.0059 & 0.0027 & 0.5340\tabularnewline
\hline 
MP min & 6 & 120 & 1 & 0.6699 & 0.0002 & 0.0883 & 0.7585\tabularnewline
\hline 
MP min & 40 & 120 & 1 & 0.0606 & 0.0002 & 0.0881 & 0.1489\tabularnewline
\hline 
MP blur & 0.01 & 0.2 & 1 & 0.9958 & 0.0042 & 0.0137 & 1.0137\tabularnewline
\hline 
MP blur & 0.01 & 0.2 & 3 & 0.5487 & 0.0100 & 0.009 & 0.5680\tabularnewline
\hline 
MP blur & 6 & 120 & 1 & 0.7240 & 0.0130 & 0.0927 & 0.8297\tabularnewline
\hline 
MP blur & 40 & 120 & 1 & 0.191 & 0.0009 & 0.0760 & 0.2679\tabularnewline
\hline 
ours & 0.01 & 0.2 & - & 0.0203 & 0.0027 & \textbf{0.0003} & 0.0234\tabularnewline
\hline 
ours & 6 & 120 & - & 0.0004 & \textbf{0.0001} & 0.0286 & 0.0291\tabularnewline
\hline 
ours & 40 & 120 & - & \textbf{0.0001} & \textbf{0.0001} & 0.0174 & \textbf{0.0176}\tabularnewline
\hline 
\end{tabular}%
\begin{tabular}{|c|}
\hline 
\multicolumn{1}{|c|}{GT localization results}\tabularnewline
\hline 
\hline 
CCs hit rate (\%) \tabularnewline
\hline 
10.5\tabularnewline
\hline 
12.6\tabularnewline
\hline 
23.9\tabularnewline
\hline 
27.0\tabularnewline
\hline 
0.10\tabularnewline
\hline 
0.11\tabularnewline
\hline 
10.4\tabularnewline
\hline 
11.9\tabularnewline
\hline 
13.9\tabularnewline
\hline 
\textbf{29.1}\tabularnewline
\hline 
\textbf{38.2}\tabularnewline
\hline 
\end{tabular}
\par\end{centering}
\caption{\label{tab:MP-min-=00003D}MP min = Meaningful Perturbation when using
a reference deletion image containing a constant value of the minimal
value in the original image. MP blur = Meaningful Perturbation when
using a reference deletion image containing a blurred version of the
original image. }
\end{table*}


\subsection{\label{subsec:Other-methods}Other methods}

In addition to Fong and Vedaldi \cite{fong2017interpretable} we tried
to get performance results for Rey-de-Castro and Rabitz \cite{rey2018targeted}.
However, no matter what hyper parameters we tried, we could not make
it generate meaningful explanations. We either got an almost full
or completely empty mask. We believe that it's due to the relatively
small capacity of the perturbation generator architecture and its
lack of ability to capture complex objects due to a too local context.
We believe that it can be seen, on Rey-de-Castro and Rabitz \cite{rey2018targeted}
visualizations, where it appears clear that the perturbation generator,
due to lack of capacity, resorted into distorting most edges in the
image, many times regardless of correlation with the relevant GT object.
This can be seen, for example, in figure 3 in \cite{rey2018targeted}
where non-cat-related edges are highlighted in addition to the cat-related
edges. However, on medical images the images are rich with edges,
and the vast majority of them are irrelevant to potential malignancy.
We tried to increase the model capacity by increasing the number of
layers, and the number of input/output channels, excluding the first
and the last layer which consist of a single output channel, but it
did not help the model converge. We tried layers numbers ranging between
1 to 30, and convolution filters number ranging from 1 to 128 on each
layer. It is possible that a larger model and/or a different architecture
for the perturbation generator will be able to converge, however,
we did not manage to find it.


\section{\label{sec:Difference-from-existing}Difference from existing methods}

In this section we will compare the proposed method with some of the
state-of-the-art methods in the field. 

In \cite{fong2017interpretable}, Fong and Vedaldi describe a loss
function which is similar to ours (namely, classification, smoothness
and sparsity approximation). However there are several main differences
between the methods:
\begin{enumerate}
\item The requirement to provide a ``deletion'' image. For example, a
blurred version of the input. It is not always clear what is the best
way to delete information, especially in domains outside natural images.
Our proposed method does not require providing any reference ``deletion''
image. A dark or blurry region in medical imaging setting does not
always equal a lack of evidence or objects. Additionally, since in
practice in their described method, an element-wise cross-fade is
performed between the original input and the reference ``deletion''
image, the possible pixel values are limited. As an extreme example,
if both the input and the reference ``deletion'' image have the
same value, their method will not be able to provide explanation that
perturb this element at all. Furthermore, non visual features like,
for example, clinical data records, or stocks market values, may have
complex relations between them, and it is not clear what the reference
``deletion'' values should be.
\item In some settings it is better to provide explanations in the original
input full resolution. An example is calcifications (accumulation
of calcium salts in a body tissue) as seen on xray, which may be as
small as a single pixel, but provide valuable medical evidence. Our
proposed method converges well and does not seem to overfit, unlike
what Fong and Vedaldi describe.
\end{enumerate}
In \cite{chang2017interpreting}, Chang \etal introduce a different
way to suppress the information within a region of an image. They
train a generative model (Variational Auto-Encoder) to impute information
inside a mask. Training a generative model (such as GAN or VAE) is
a non trivial matter, and our proposed method does not require it.

Rey-de-Castro and Rabitz in \cite{rey2018targeted} propose to train
an additional model, to generate image perturbations. While there
are several similarities to our proposed methods, there are key differences:
\begin{enumerate}
\item The method requires an additional perturbation generator model. Firstly,
choosing an appropriate architecture may prove difficult. On one hand,
a too simplistic model may not be sufficient to capture the desired
behavior, due to low capacity and/or lack of large enough context.
On the other hand, a complex model may be difficult to train. In contrast,
our proposed method does not require constructing any additional model
architecture, as we perturb the values of the input directly. It's
worth noting that our modification of the input is non-linear, as
it originates from gradients propagation through a non-linear function
(the original model). 
\item The lack of total variation in the proposed loss function does not
encourage ``simple`` explanations.
\item Rey-de-Castro and Rabitz optimize for L1, while we optimize to minimize
a closer approximation of L0 (eq. \ref{eq:6}). 
\end{enumerate}
Additionally, as we describe in section \ref{subsec:Other-methods}
it proved difficult, in practice, to make the perturbation generator
provide meaningful explanations that go beyond perturbing most edges
in the image, regardless of their relation to any object.

\section{\label{sec:Conclusions}Conclusions}

In this paper we presented a new method for explaining the decision
of a given model on a specified input. The method is based on adversarial
examples generation, which, when constrained to simple changes, is
shown to provide well-localized meaningful explanations. We test the
method on a public dataset of breast mammogram images, and show that
it significantly outperforms the current state-of-the art in the field,
both quantitatively, using heuristic metrics and ground-truth-based
comparison, and qualitatively.

While the analyzed network is never exposed to localization information,
the proposed explanation method also extracts meaningful local cues.
Extending this functionality within the framework of weakly-supervised
segmentation is part of currently ongoing work.

{\small{}\bibliographystyle{ieee}
\nocite{*}
\bibliography{egpaper_for_arxiv}

\begin{thebibliography}{10}\itemsep=-1pt

\bibitem{chang2017interpreting}
C.-H. Chang, E.~Creager, A.~Goldenberg, and D.~Duvenaud.
\newblock Interpreting neural network classifications with variational dropout
  saliency maps.
\newblock In {\em Proc. NIPS}, 2017.

\bibitem{dabkowski2017real}
P.~Dabkowski and Y.~Gal.
\newblock Real time image saliency for black box classifiers.
\newblock In {\em Advances in Neural Information Processing Systems}, pages
  6967--6976, 2017.

\bibitem{fong2017interpretable}
R.~C. Fong and A.~Vedaldi.
\newblock Interpretable explanations of black boxes by meaningful perturbation.
\newblock {\em arXiv preprint arXiv:1704.03296}, 2017.

\bibitem{goodfellow2018explaining}
I.~J. Goodfellow, J.~Shlens, and C.~Szegedy.
\newblock Explaining and harnessing adversarial examples. arxiv, 2018.

\bibitem{heath2000digital}
M.~Heath, K.~Bowyer, D.~Kopans, R.~Moore, and W.~P. Kegelmeyer.
\newblock The digital database for screening mammography.
\newblock In {\em Proceedings of the 5th international workshop on digital
  mammography}, pages 212--218. Medical Physics Publishing, 2000.

\bibitem{kingma2014adam}
D.~P. Kingma and J.~Ba.
\newblock Adam: A method for stochastic optimization.
\newblock {\em arXiv preprint arXiv:1412.6980}, 2014.

\bibitem{kurakin2016adversarial}
A.~Kurakin, I.~Goodfellow, and S.~Bengio.
\newblock Adversarial examples in the physical world.
\newblock {\em arXiv preprint arXiv:1607.02533}, 2016.

\bibitem{modas2018sparsefool}
A.~Modas, S.-M. Moosavi-Dezfooli, and P.~Frossard.
\newblock Sparsefool: a few pixels make a big difference.
\newblock {\em arXiv preprint arXiv:1811.02248}, 2018.

\bibitem{rey2018targeted}
R.~Rey-de Castro and H.~Rabitz.
\newblock Targeted nonlinear adversarial perturbations in images and videos.
\newblock {\em arXiv preprint arXiv:1809.00958}, 2018.

\bibitem{robbins1951}
H.~Robbins and S.~Monro.
\newblock A stochastic approximation method.
\newblock {\em Ann. Math. Statist.}, 22(3):400--407, 09 1951.

\bibitem{simonyan2013deep}
K.~Simonyan, A.~Vedaldi, and A.~Zisserman.
\newblock Deep inside convolutional networks: Visualising image classification
  models and saliency maps.
\newblock {\em arXiv preprint arXiv:1312.6034}, 2013.

\bibitem{szegedy2016rethinking}
C.~Szegedy, V.~Vanhoucke, S.~Ioffe, J.~Shlens, and Z.~Wojna.
\newblock Rethinking the inception architecture for computer vision.
\newblock In {\em Proceedings of the IEEE conference on computer vision and
  pattern recognition}, pages 2818--2826, 2016.

\bibitem{zeiler2012adadelta}
M.~D. Zeiler.
\newblock Adadelta: an adaptive learning rate method.
\newblock {\em arXiv preprint arXiv:1212.5701}, 2012.

\bibitem{zhang2018detecting}
C.~Zhang, Z.~Yang, and Z.~Ye.
\newblock Detecting adversarial perturbations with saliency.
\newblock {\em arXiv preprint arXiv:1803.08773}, 2018.

\end{thebibliography}
 }{\small \par}
\end{document}